\documentclass[11pt,a4paper]{article}
\usepackage{emnlp2020}
\usepackage{times}
\usepackage{latexsym}
\usepackage{subfigure}

\usepackage{microtype}

\aclfinalcopy 

\usepackage{algorithm}
\usepackage[noend]{algpseudocode}

\makeatletter
\renewcommand{\ALG@beginalgorithmic}{\tiny}
\makeatother

\algrenewcommand\alglinenumber[1]{\tiny #1:}
 \usepackage{amssymb}
 \usepackage{array}
\usepackage{multirow}
\usepackage{soul}
\usepackage{xcolor}
\usepackage{graphicx}

\title{Adversarial Black-Box Attacks On Text Classifiers Using Multi-Objective Genetic Optimization Guided By Deep Networks}

\author{Alex Mathai\textsuperscript{1}, Shreya Khare\textsuperscript{1}, Srikanth Tamilselvam\textsuperscript{1}, Senthil Mani\textsuperscript{1} \\ \\
  \textsuperscript{1}IBM Research Labs \\ \\
  \texttt{ \small{\{alex.mathai1, skhare34, srikanth.tamilselvam, sentmani\}@in.ibm.com }}}

\date{}

\begin{document}
\maketitle
\begin{abstract}

We propose a novel genetic-algorithm technique that generates black-box adversarial examples which successfully fool neural network based text classifiers. We perform a genetic search with multi-objective optimization guided by deep learning based inferences and Seq2Seq mutation to generate semantically similar but imperceptible adversaries. We compare our approach with DeepWordBug (DWB) on SST and IMDB sentiment datasets by attacking three trained models viz. char-LSTM, word-LSTM and elmo-LSTM. On an average, we achieve an attack success rate of $65.67\%$ for SST and $36.45\%$ for IMDB across the three models showing an improvement of $49.48\%$ and $101\%$ respectively. Furthermore, our qualitative study indicates that $94\%$ of the time, the users were not able to distinguish between an original and adversarial sample. 
\end{abstract}
\section{Introduction}
Deep Neural Networks (DNNs) have witnessed tremendous success in day to day applications like chat-bots and self-driving vehicles. This has made it imperative to test the robustness of these models prior to their deployment for public consumption. \citet{Szegedy} first demonstrated the vulnerability of DNNs in computer vision by strategically fabricating adversarial examples.  
These carefully curated examples remain correctly classified by a human observer but can fool a targeted model.

Adversarial generation can be broadly classified into two types based on their access to the model. The generation algorithms that rely on the model internals are known as white-box algorithms \cite{papernot2017practical,kurakin2016adversarial,baluja2017adversarial}. Others \cite{su2019one,sarkar2017upset,cisse2017houdini} which work without any knowledge about the model parameters and gradients are known as black box algorithms and are more prevalent in real-world scenarios. These works triggered a flurry of research towards i) evaluating the vulnerability of DNNs by attacking them with unperceivable perturbations, ii) measuring sensitivity and iii) developing defense mechanisms against such attacks. As industry increasingly employs DNN models in NLP tasks like text classification, sentiment analysis and question answering, such systems are also under the threat of adversaries \cite{zhao2019gender,horton2016microsoft}. 



 Existing adversarial techniques used to misguide image-based systems cannot be applied to text based systems because of two key reasons. First, image pixel values are continuous whereas words are discrete tokens. Hence, small changes in pixel values go unnoticed but the same cannot be said for typos or misfit words. 
 Second, due to its discrete nature, input words are mapped to a continuous space using word embeddings. Therefore, any  gradient computation is done only on the embeddings and not directly on the input words. Hence, previous gradient based attacks like FGSM \cite{Goodfellow} are difficult to apply on text models. 
 In addition to the above challenges, perturbations applied to original text should be carefully crafted so that the adversarial sample is still a) \textit{structurally similar}: the user should not be able to find differences in a quick glance, b) \textit{semantically relevant}: the generated sample should bear the same meaning as the source sample and c) \emph{grammatically coherent}: the generated sample should look natural and be grammatically correct.

To generate text adversaries satisfying multiple criteria, we formulate a multi-objective optimization problem, that once solved, results in the generation of text samples with adversarial properties. Thus, we model adversarial generation as a natural search space selection problem where the best adversaries are selected from a pool of samples initialized by different techniques. 
This selection is based on multiple objectives, where each objective measures a desirable adversarial quality. The selected samples then further undergo mixing (reproduction) and slight modifications (mutation) to create even better adversaries. As this model resembles the biological process of evolution, we rely on genetic algorithms as a tool to solve this optimization problem. Our genetic algorithm works in tandem with a sophisticated SeqtoSeq model for mutation, embedding based word replacements, textual noise insertions and deep learning based objectives to ensure a good selection of candidate solutions.

In summary, we propose a novel multi-objective genetic optimization that works in tandem with deep learning models to generate text adversaries satisfying multiple criterion. We conduct an extensive quantitative and qualitative study to measure the quality of the text adversaries generated and compare them with DeepWordBug (DWB) \cite{deepwordbug}.
\section{Existing Literature}

Most black-box adversarial algorithms for text have been rule based or have used supervised techniques. Such attacks are generally applied at sentence, word and character granularities.

At the sentence granularity, \citet{jia2017adversarial} shows how the addition of an out-of-context sentence acts like an adversary for comprehension systems. However, these sentences are easily identifiable as out-of-context by human evaluators.

At the word granularity, perturbations are generally introduced by selective word replacements. \citet{papernot2017practical} focuses on minimum word replacements for generating adversarial sentences. \citet{alzantot2018generating} uses single-objective genetic algorithms to generate adversaries by replacing words with their synonyms at an attempt to preserve semantics. \citet{sato2018interpretable} introduces direction vectors which associate the perturbed embedding to a valid word embedding from the vocabulary.

To target character based models, adversarial attacks focusing at a character level have also surfaced. \citet{gao2018black} suggests a method to introduce random character level insertions using nearest keyboard neighbours. \citet{li2018textbugger} focuses on attacking with character and word level perturbations. They introduce misspellings which map to unknown tokens when the model refers to a dictionary. However, introducing numerous misspellings makes it very difficult to comprehend adversaries.

Multi-objective optimization has been effective for many applications like image-segmentation and generating speech adversaries for ASR systems \cite{article,Khare2018AdversarialBA}. 
Our multi-objective optimization algorithm leverages the use of both selective word replacements and character level error insertion strategies. But unlike other works that rely on simple heuristics, we ensure that the modifications we make are essential and minimal by infusing guidance from deep learning based objectives. Guided by  networks like Infersent \cite{DBLP:journals/corr/ConneauKSBB17}, OpenAI GPT \cite{Radford2018ImprovingLU} and SeqtoSeq  \cite{bahdanau2014neural}, we create adversaries that are structurally similar, semantically relevant and grammatically coherent.


\section{Proposed Approach}
Multi-objective optimization is the process of optimizing multiple objective functions \cite{Coello:2006:EAS:1215640}. 
The interaction among different objectives gives rise to a set of compromised solutions known as \emph{Pareto-optimal} solutions \cite{pareto}. Evolutionary algorithms are one class of popular approaches to generate such \emph{Pareto-optimal} solutions.

Inspired from evolutionary systems, we consider adversarial text generation as an evolutionary process to generate adversarial samples satisfying multiple criteria.  Carefully designed deep learning objectives and text processing algorithms assist in creating improved candidate solutions (text-samples) for fooling the classifier.
Further, we propose various metrics to assess the fitness of a candidate (text-sample) within a population.
After multiple generations, we finally apply heuristics to pick the most appropriate adversary that satisfies the desired adversarial properties discussed in the introduction.

An overview of the proposed algorithm for producing adversarial text samples for a given input is shown in Algorithm $1$. 
The entire process is divided into six main stages typical to an evolutionary process: i) Population Initialization, ii) Mutation, iii) Fitness Evaluation iv) Mating Pool Identification, v) Crossover and vi) Dominance Evaluation.

\begin{figure*}
\centering
\includegraphics[height=6.5cm,width=16cm]{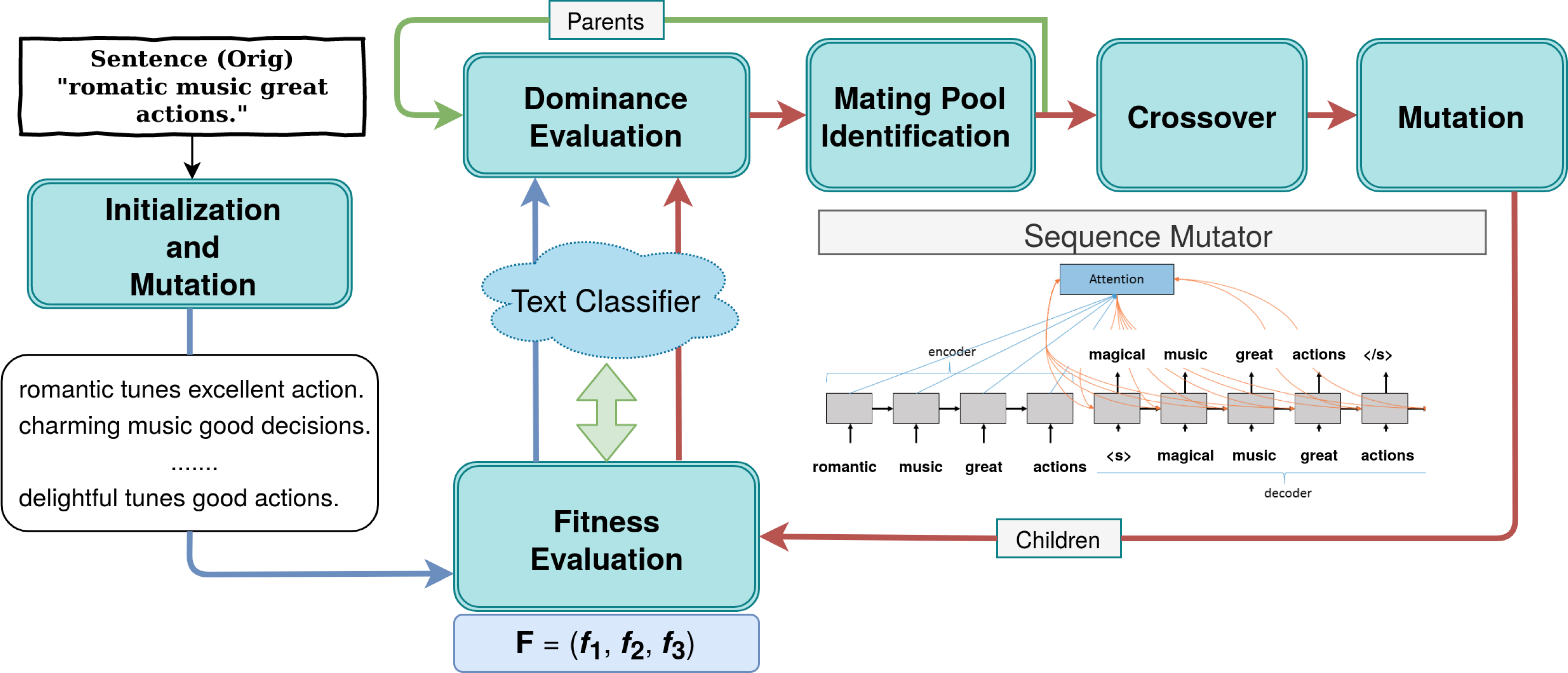}
\caption{Adversarial Generation System using Multi-objective Genetic Algorithm}
\label{fig:sys_arch}
\end{figure*}

Our algorithm supports two ways of fooling classifiers. 1) Selectively substituting words with semantically equivalent terms. 2) Combining substitution of selective words along with introduction of typographical errors. We refer to these two modes of attacks as $Attack_{Single}$ and $Attack_{Combined}$ respectively. We report the results for both the types in the experiments section. The flow of the algorithm across the $6$ stages can be seen in Fig. \ref{fig:sys_arch} and lines $65$ to $94$ in Algorithm $1$. The details of each stage are explained in the following sections.
\begin{algorithm}[htbp]
\caption{Algorithm for generating an adversarial text sample, given a reference input sample}

\begin{algorithmic}[1]

\State $t_{Orig} \gets$ Sample text,$w_{cf} \gets$ Counter-Fitted Vectors, $popsize \gets 64$
\State $Enc$, $Dec \gets$ Encoder, Decoder of Sequence Mutator, $\delta{} \gets 0.6$
\State $Mode_{Attack} \gets$ 1 strategy from [$Attack_{Single}$, $Attack_{Combined}$]\\

\Procedure{Mutation}{$Sents,Mode_{Mutation}$}
\State \% Performs mutation in a Sentence \% 
\If { $Mode_{Mutation}$ is $Typo$ }
\State \% Add Typo in a Sentence \% 
\State $muts \gets AddTypo(Sents)$
\Else
\State \% Pass Sentence through Sequence Mutator \%
\State $muts \gets Dec(Enc(Sents))$
\EndIf\\
\Return $muts$
\EndProcedure\\

\Procedure{Initialization}{$t_{Orig},Mode_{Mutation},popsize$}
\State \% Initialize the Population \%
\State $P \gets list[]$
\State Find important words $w_{imp}$ in $t_{Orig}$ using $Infer$
\State Find substitute words $w_{sub}$ for the $w_{imp}$ with nearest neighbour search
\State Allow only synonyms from $w_{sub}$ using $w_{cf}$ to ensure $similarity > \delta{}$ \\

\State \%  Find all combinations of substitutions and create sentences \%
\State $combinations \gets powerset(w_{sub})$
\For {$w \in combinations $}
\State Apply $w$ in $t_{Orig}$ and add to $P$
\EndFor
\State $Pop \gets popsize$ samples from $P$
\State $Pop \gets Mutation(P,Mode_{Mutation})$\\
\Return $Pop$
\EndProcedure \\

\Procedure{Fitness Evaluation}{$Population$}
\State $set \gets \{\}$
\State Store fitness scores for $Population$ using Eqn. (\ref{eqn:eq1}) to (\ref{eqn:eq6}) in $set$.\\
\Return $set$
\EndProcedure \\

\Procedure{GetDominating}{$Population$, $popsize$}
\State \%  Use the NSGA-2 Algorithm to select from Parents and Children \%
\State $P \gets SelectNSGA2(Population, popsize)$\\
\Return $P$
\EndProcedure \\

\Procedure{Mating Pool}{$Parents, Fitness$}
\State \%  List of Fittest Parent Pairs for Reproduction \%
\State $pool \gets TournamentSelection(Parents, Fitness)$\\
\Return $pool$
\EndProcedure \\

\Procedure{Crossover}{$ParentPairs$}
\State $set \gets \{\}$ 
\For {$p1,p2$ in $ParentPairs$}:
    \State \%  Perform Crossover amongst normal forms \%
    \State $c1, c2 \gets SinglePoint(p1.Ind,p2.Ind) $ \\
    \State \%  Perform Crossover amongst mutated forms \%
    \State $c3, c4 \gets SinglePoint(p1.IndMut,p2.IndMut) $
    \State $set \gets set \cup \{c1,c2,c3,c4\} $
\EndFor\\
\Return $set$
\EndProcedure \\

\Procedure{GA}{$Population_{Old},Fitness_{Old},Mode_{Mutation}$}
\State \%  Performs 1 iteration of the Genetic Algorithm \%
\State $parents 
\gets  MatingPool(Population_{Old},Fitness_{Old})$
\State $children \gets Crossover(parents)$
\State $children \gets Mutation(children,Mode_{Mutation})$
\State $Population_{New} \gets GetDominating(parents \cup children)$
\State $Fitness_{New} \gets FitnessEvaluation(Population_{New})$\\
\Return $Population_{New}$,$Fitness_{New}$
\EndProcedure \\

\Procedure{GA-Execute}{$t_{Orig},popsize,Mode_{Attack}$}
\If {$Mode_{Attack}$ is $Attack_{Single}$}
\State \%  Complete population has only word swaps \%
\State ${P_{S}}_{i} \gets Initialization(t_{Orig},Swaps,popsize)$
\State ${F_{S}}_{i} \gets FitnessEvaluation({P_{S}}_{i})$
\Else
\State \%  Half population \textbf{($P_{S}$)} has only word \textbf{swaps} \%
\State ${P_{S}}_{i} \gets Initialization(t_{Orig},Swaps,popsize//2)$
\State ${F_{S}}_{i} \gets FitnessEvaluation({P_{S}}_{i})$ \\
\State \%  Half population \textbf{($P_{T}$)} has word swaps with \textbf{typos} \%
\State ${P_{T}}_{i} \gets Initialization(t_{Orig},Typos,popsize//2)$
\State ${F_{T}}_{i} \gets FitnessEvaluation({P_{T}}_{i})$
\EndIf\\

\State \%  Perform GA multiple times \%
\Repeat
\If {$Mode$ is $Attack_{Single}$}
\State \%  Perform an iteration of GA on the whole population \%
\State ${P_{S}}_{i+1},{F_{S}}_{i+1} \gets  GA({P_{S}}_{i},{F_{S}}_{i},Swaps)$
\State ${P_{Total}}_{i+1} = {P_{S}}_{i+1}$
\Else
\State \%  Perform an iteration of GA on the word \textbf{swaps} population \%
\State ${P_{S}}_{i+1},{F_{S}}_{i+1} \gets  GA({P_{S}}_{i},{F_{S}}_{i},Swaps)$ \\
\State \%  Perform 1 iteration of GA on the \textbf{typo} population \%
\State ${P_{T}}_{i+1},{F_{T}}_{i+1} \gets  GA({P_{T}}_{i},{F_{T}}_{i},Typos)$
\State ${P_{Total}}_{i+1} = {P_{S}}_{i+1} \cup {P_{T}}_{i+1}$
\EndIf
\Until{ ${P_{Total}}_{i+1} == {P_{Total}}_{i}$ or $i > maxiters$ }\\
\Return ${P_{Total}}_{i+1}$
\EndProcedure

\label{algo:1}
\end{algorithmic}
\end{algorithm}



\subsection{Initialization} 
In this phase (Algorithm $1$, Lines $15$-$28$), candidate samples in the population are initialized. For each sample, we lowercase and remove punctuation from the original text and perform word-level perturbations. These perturbations are done by first identifying the most important words in the text and then replacing them with semantically equivalent terms. To find the important words in the text, we max-pool the penultimate layer of the InferSent Model, which has been trained as a universal sentence encoder. Once the most important words are identified, we search for their nearest neighbours in the GloVe embedding space \cite{pennington-etal-2014-glove}. This is done by finding similar vectors having cosine similarity more than $\delta{}$ (a hyper-parameter).
We leverage the FAISS toolkit \cite{Johnson2017BillionscaleSS} for this step. To make sure that these neighbours are synonyms, we also verify that the cosine similarity is more than $\delta{}$ even in the counter-fitted vector space \cite{mrksic-etal-2016-counter}.

After finding the possible substitutions, we then initialize a population of candidate samples with all the possible combinations of original and substituted words. From this population, a small subset is selected randomly to make the genetic computation tractable. This careful initialization maintains diversity in the population and seeding it with original text guarantees faster convergence.

\subsection{Mutation}
The diversity of the existing population is further enhanced by mutation. In the text space, the mutation of a candidate is defined as a suitable word substitution. We rely on $2$ main approaches for substitutions 1) \emph{GloVe Mutator} - a nearest neighbour search in the counter-fitted vector space as mentioned in the initialization phase and 2) \textit{Sequence Mutator} - an encoder-decoder deep learning model to perform mutations in a sentence. 

The \emph{Sequence Mutator} is trained as a sequence to sequence bi-directional LSTM model with attention for the sole objective of predicting the same input that is fed to it. Hence the \textit{Sequence Mutator} is made to iteratively predict a word given its left and right context in the input sentence until all the words in the input sentence are predicted.

This training objective makes the \textit{Sequence Mutator} very similar to a language model. 
However, a key difference between the two, is the initial state supplied to the decoder.  In a language model, the initial state is generally a vector of zeros, however, in the \textit{Sequence Mutator} it is the context vector passed by the encoder. This context vector helps in copying the input sentence effectively. We perform beam search to get the top $5$ predictions from the \textit{Sequence Mutator}. This step helps to randomly sample a mutated form of the input sentence. 

In general, it is observed that the most probable prediction made by the \textit{Sequence Mutator} is the original sentence itself and the remaining $4$ predictions are similar sentences with $1$ or $2$ words substituted probabilistically during decoding. We use the fairseq toolkit \cite{ott2019fairseq} for fast prototyping of sequence to sequence model architectures.

It is very important to note that in the case of $Attack_{Single}$, we pass the entire population through the \textit{Sequence Mutator} or \emph{Glove Mutator}, whereas, in the case of $Attack_{Combined}$, we pass half of the population through the \textit{Sequence Mutator} or \emph{Glove Mutator} and for the other half we introduce typographical errors (Algorithm $1$, Lines $66$-$77$).

For typographical errors, we randomly select one or multiple words wherein the maximum number of words selected are less than $10$\% of the length of the sentence. For the selected words, only $1$ of the following ways of typographical error introduction is chosen. i) Character swaps \cite{soni2019} and ii) QWERTY character map exchanges \cite{pruthi2019combating}. Character swaps are implemented by randomly choosing a character from the word and then swapping its place with an adjacent character. For the QWERTY character map exchange, the first step is to maintain a list of the adjacent symbols for every character in the QWERTY keyboard. When a character is randomly chosen, it is replaced by any one of the adjacent symbols. Both errors are included keeping in mind that they constitute a large portion of the most common typographical errors made while typing.

These steps are outlined in Algorithm $1$ (Lines $5$-$13$). To ensure diversity, we maintain both the normal and mutated forms of each candidate in the population. In Algorithm $1$, the normal and mutated forms of a candidate individual are denoted as $Ind$ and $IndMut$ respectively.

\subsection{Fitness Evaluation} 
\begin{figure}[h]
\centering
\includegraphics[width=0.45\textwidth]{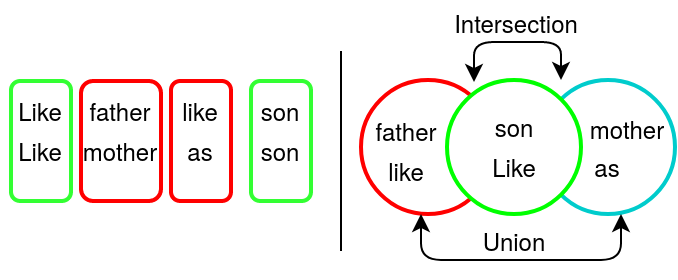}
\caption{Positional Jaccard}
\label{fig:jaccard}
\end{figure}

In-order to generate better candidate samples in the next generation, we need to identify the fittest candidates from the current population. The fitness function (Algorithm $1$, Lines: $30$-$33$) evaluates the adversarial nature of the samples i.e. the \textit{structural similarity}, the \textit{semantic relevance} and the ability to fool the target classification model. 
The fitness function for each candidate is a $3$ element vector $\vec{F}$.
\begin{equation}
\label{eqn:eq1}
    \vec{F} = (f1,f2,f3)
\end{equation}
The first objective value $f1$ measures the absolute change in the posterior probabilities when an individual candidate ($Ind$) and when the original text ($Orig$) are fed to the classifier ($Model$).
\begin{equation}
\label{eqn:eq2}
    f1 = abs(Model(Ind) - Model(Orig))
\end{equation}
The second objective value measures the structural similarity between $Ind$ and $Orig$. This is calculated by finding the positional jaccard co-efficient between the $2$ sentences. As shown in Fig. \ref{fig:jaccard}, the intersection is $2$ words and the union is $6$ words. Hence, $PosJaccard = 1/3$.
\begin{equation}
\label{eqn:eq4}
    PosJaccard(A,B) = |A|\cap|B|/|A|\cup|B|
\end{equation}
\begin{equation}
\label{eqn:eq5}
    f2 = PosJaccard(Ind,Orig)
\end{equation}
The third and final objective value measures the semantic relevance between $Ind$ and $Orig$. For this, we use the InferSent  model to encode the sentences to fixed size vectors after which the cosine similarity is calculated.
\begin{equation}
\label{eqn:eq6}
    f3 = Sim(\vec{Infer(Orig)},\vec{Infer(Ind)}) 
\end{equation}
In this manner, we incorporate each adversarial quality as an objective value derived from a deep learning model. By doing so, we ensure that only adversaries that have high structural similarity and semantic relevance progress to later iterations and form a major portion of the gene pool. 

\subsection{Mating Pool Identification} Once we calculate all the fitness functions, we then identify the fittest parents from the population. This step (Algorithm $1$, Lines: $40$-$43$) conducts a tournament selection to choose parents that will further undergo a crossover operation to produce children. In this tournament, a candidate sample completely dominates another opponent if the sample is at least as good as the opponent across all objective values. If not, the tie breaker for the tournament is the \emph{crowding metric} which is a sub-part of the NSGA-2 \cite{Deb:2002:FEM:2221359.2221582} algorithm.

\subsection{Crossover}
To enhance diversity in the adversarial samples, crossover is performed among the parent candidates.  For crossover (Algorithm $1$, Lines: $45$-$54$) between two parents, we use a single point structural crossover. A word index is chosen randomly and then parts of the text after the index are swapped between the $2$ samples. As each candidate sample has a normal and mutated form, we perform crossovers between normal and mutated forms separately. Hence $2$ candidate samples produce $4$ offsprings - $2$ from crossover of the normal forms and $2$ from crossover of the mutated forms.

\subsection{Dominance Evaluation}
After the creation of children we proceed to evaluate their objective values. Post this, we are left with two sets of populations - the first is the set of parents and the second is the set of children. To maintain a constant population size during adversarial generation, we need to choose the best candidates from the combined group containing both parents and children. For this (Algorithm $1$, Lines: $35$-$38$), we employ our selection algorithm based on NSGA-2 \cite{Deb:2002:FEM:2221359.2221582}.  In its standard formulation NSGA-2 groups candidates together if and only if every candidate in the group is not better than every other candidate in that group across all the objective values. Within the same group, ranks are given to candidates based on \emph{crowding metrics}. We find that on experimentation, the standard formulation is successful at selecting good quality adversaries from the population.

\begin{table*}[t]
\small
\centering
\resizebox{\textwidth}{!}{%

\begin{tabular}{|m{12cm}|m{1.2cm}|m{1cm}|}
\hline
\multicolumn{1}{|c|}{Generated Adversarial Examples}                                                                                                                                                                      & Classifier     & Label Flip \\ \hline
 
\begin{tabular}[c]{@{}l@{}}\textbf{SST :} A \st{cumbersome} \textcolor{blue}{burdensome}  and cliche-ridden movie greased with every emotional \\ device known to man.
\end{tabular}                                                           & word-LSTM & - to +   \\ \hline
\begin{tabular}[c]{@{}l@{}}\textbf{SST :} Birot is a competent enough filmmaker but her \st{story} \textcolor{blue}{tale} has nothing fresh or very \\ exciting about it.
\end{tabular}                                        & word-LSTM & - to + \\ \hline

\textbf{SST :} A taut, \st{intelligent} \textcolor{blue}{sensible} psychological drama. & char-LSTM & + to -   \\ \hline


\begin{tabular}[c]{@{}l@{}}\textbf{SST :} Has it ever been possible to say that Williams has \st{truly} \textcolor{blue}{genuinely} inhabited a character?\end{tabular}                                                                           & char-LSTM & - to + \\ \hline



\begin{tabular}[c]{@{}l@{}}\textbf{SST :} Like most Bond outings in recent years some of the stunts  are so  \st{outlandish} \textcolor{blue}{outrageous} \\ that they border on being cartoon-like.\end{tabular}                                    & elmo-LSTM & - to + \\ \hline


\begin{tabular}[c]{@{}l@{}}\textbf{IMDB :} 
caught the tail ... dont see the shooting only adds to the chillness of the plot to see both \\ \st{child}  \textcolor{blue}{infant} and adult alike   struggle  to  \st{comprehend} \textcolor{blue}{scavenge}  and come to terms   with the \\ \st{senseless} \textcolor{blue}{inane} shootings  was at times \st{overwhelming}  \textcolor{blue}{surprising} and will admit that i shed quite.

\\\end{tabular}                                                                                                                                                       & elmo-LSTM & + to - \\ \hline
\end{tabular}
}
 \caption{Adversaries generated using our algorithm for different classifiers. The blue text indicates perturbations.}
 \label{table:sst_samples}
\end{table*}
\subsection{Final Selection}
This process of evolution is repeated until the last iteration ($60$ steps) is reached or until population convergence (Algorithm $1$, Lines: $79$-$93$). To measure the grammatical coherence of the candidate samples, we pass the adversaries through the OpenAI GPT language model (\textit{LM}) \cite{Radford2018ImprovingLU} and calculate the word normalized perplexity loss for each of the adversaries. The lower the loss for a particular candidate sample the better its quality. In order to select the best adversarial candidate from the final population, it was essential to derive a metric that emphasized on the desired adversarial qualities. For a given candidate individual ($Ind_{i}$), we would prefer that it has higher structural similarity ($f2_{i}$) and higher semantic relevance ($f3_{i}$) while simultaneously having a lower word normalized perplexity ($LM_{i}$).
\begin{equation}
\label{eqn:eqn11}
    score_{i} = (f2_{i}*f3_{i})/(LM_{i})^{\alpha}
\end{equation}
Using this intuition, the metric ($score_{i}$ with $\alpha = 1.2$) was calculated for each adversary and the best scoring adversaries were chosen from the test sets of SST-2 and for $1000$ reviews from IMDB corpus.
On a $K80$ GPU, each iteration takes $0.5$ seconds for Glove and $3$ seconds for the Sequence Mutator.

\section{Experimental Results and Analysis}

\begin{figure*}[t]
	\begin{center}
			\includegraphics[width=0.85\textwidth,height=0.37\textwidth]{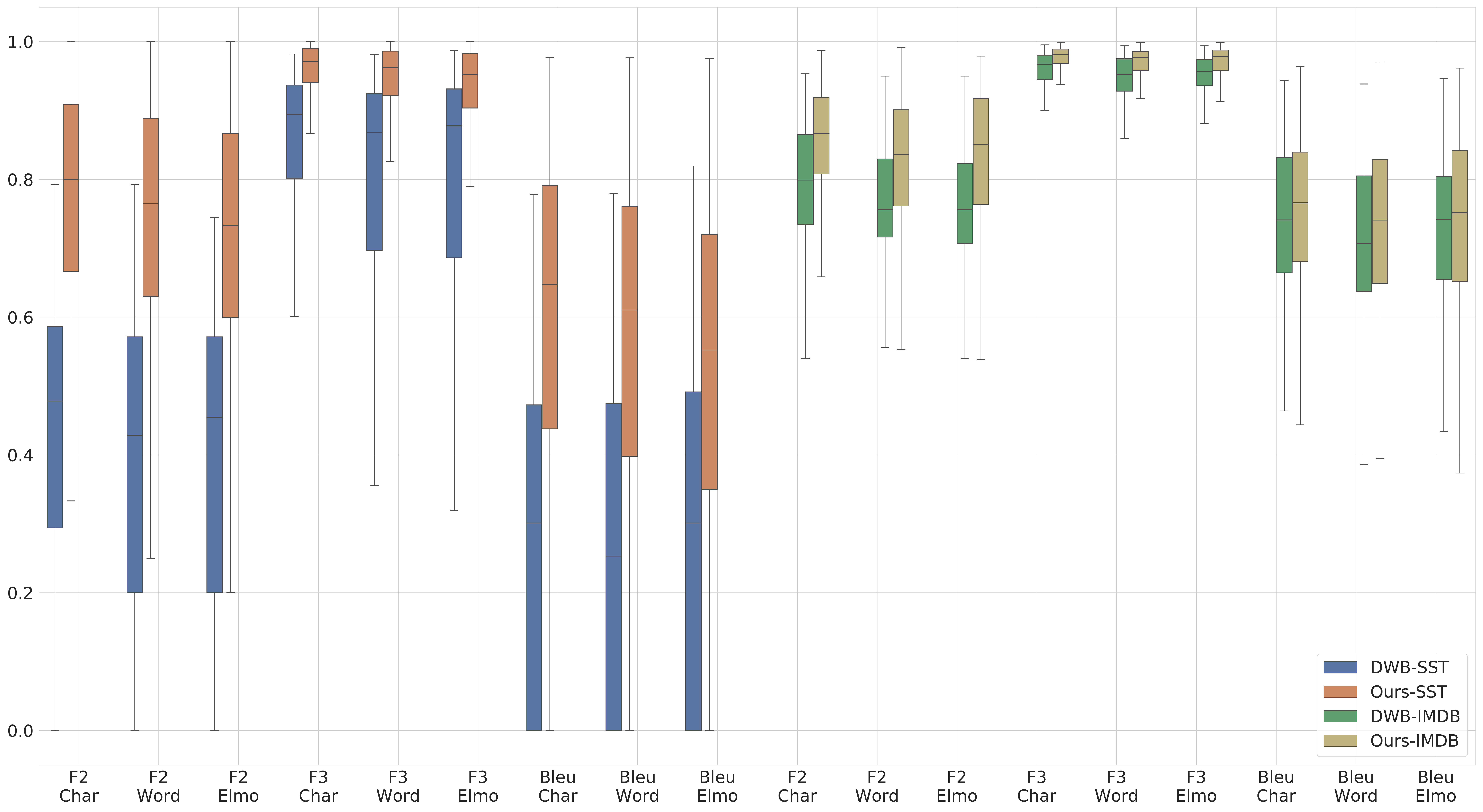}
			\includegraphics[width=0.85\textwidth,height=0.37\textwidth]{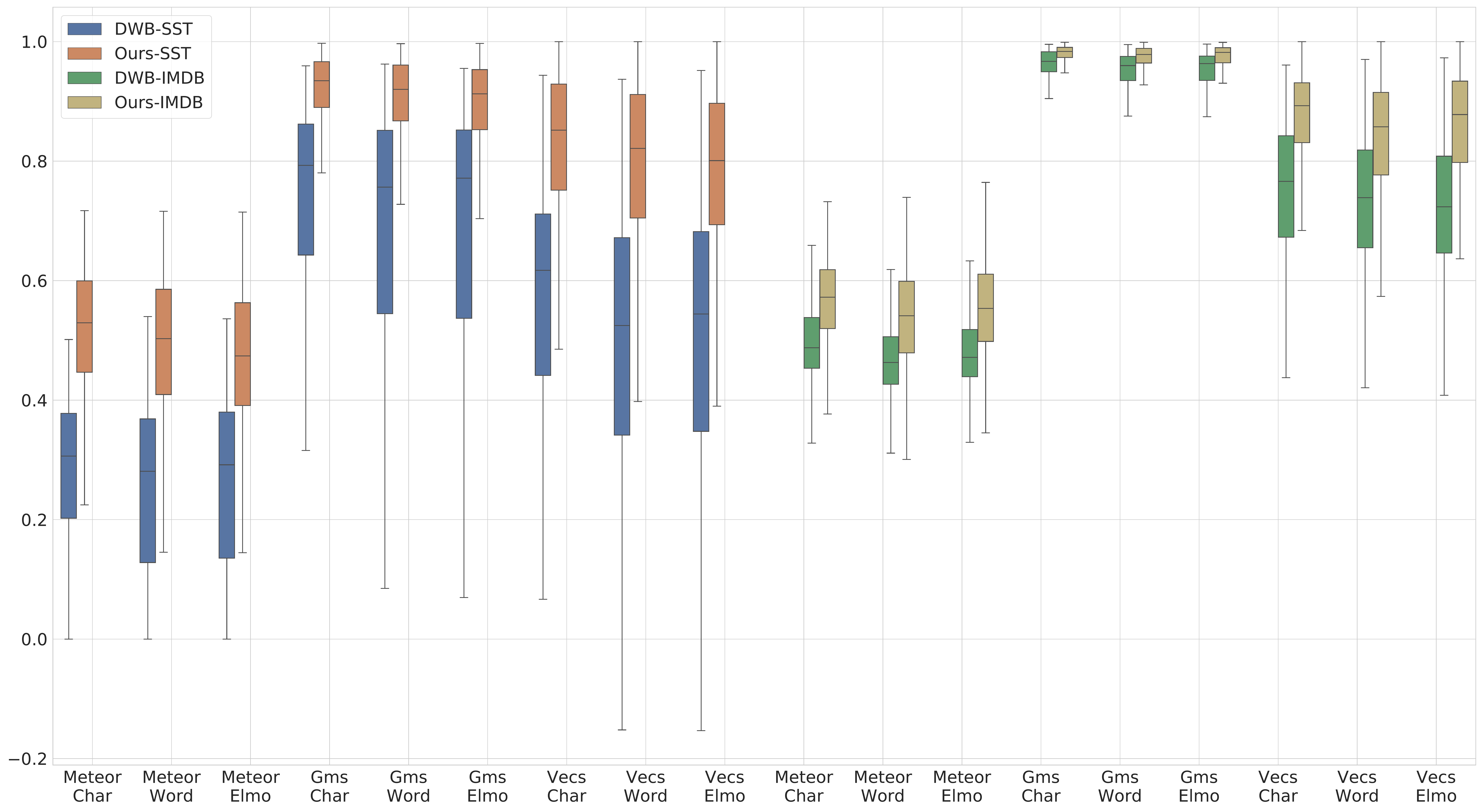}
			\label{fig:resExactTW}
		\caption{Metrics indicating semantic and structural similarity  across adversarial and original sentences}
		\label{fig:reschunks}
	\end{center}
\end{figure*}

\begin{table*}[t]
\centering
\resizebox{\textwidth}{!}{%
\begin{tabular}{|c|c|c|c|c|c|c|c|c|c|}
\hline
Dataset & Classifier & Accuracy (Original) & \multicolumn{2}{c|}{Success Rate (Glove Mutator)} & \multicolumn{2}{c|}{Success Rate (Seq2Seq Mutator)} & DWB & \multicolumn{2}{c|}{Accuracy (Degraded)} \\ \hline
- & - & - & $Attack_{Single}$ & $Attack_{Combined}$ & $Attack_{Single}$ & $Attack_{Combined}$ & (Success Rate) & Ours & DWB \\ \hline
\multirow{3}{*}{SST} & char-LSTM & 75.26\% & 52.61\% & 70.75\% & \textbf{63.68\%} & \textbf{73.67\%} & 42.29\% & \textbf{26.33\%} & 57.71\% \\ \cline{2-10} 
 & word-LSTM & 82.95\% & 37.05\% & 61.35\% & \textbf{47.4\%} & \textbf{63.52\%} & 52.21\% & \textbf{36.48\%} & 47.79\% \\ \cline{2-10} 
 & elmo-LSTM & 87.02\% & 28.20\% & 57.72\% & \textbf{39.8\%} & \textbf{59.82\%} & 37.31\% & \textbf{40.18\%} & 62.69\% \\ \hline
\multirow{3}{*}{IMDB} & char-LSTM & 87\% & 30.1\% & 37.5\% & \textbf{34.82\%} & \textbf{39.5\%} & 14.7\% & \textbf{60.5\%} & 85.3\% \\ \cline{2-10} 
 & word-LSTM & 90.4\% & 19.7\% & 41.3\% & \textbf{23.58\%} & \textbf{42.7\%} & 24.9\% & \textbf{57.3\%} & 75.1\% \\ \cline{2-10} 
 & elmo-LSTM & 92.34\% & 17.29\% & 26.19\% & \textbf{22.71\%} & \textbf{27.15\%} & 14.7\% & \textbf{72.85\%} & 85.3\% \\ \hline
\end{tabular}
}
\caption{Success rate and Model performance degradation  } 
\label{tab:my-table}
\end{table*}

\begin{table*}[t]
\centering
\resizebox{0.55\textheight}{!}{%
\begin{tabular}{|c|c|c|c|c|c|c|c|c|c|}
\hline
Dataset & Classifier & \multicolumn{2}{c|}{AWR Metric} & \multicolumn{3}{c|}{DWB} & \multicolumn{3}{c|}{ $Attack_{Combined}$ {[}Seq2Seq{]}} \\ \hline
-- & -- & Ours & DWB & char-LSTM & word-LSTM & elmo-LSTM & char-LSTM & word-LSTM & elmo-LSTM \\ \hline
\multirow{3}{*}{SST} & char-LSTM & 1.8 & 5.93 & -- &  39.59\% &  43.17\% & -- & 25.92\% & 19.55\% \\ \cline{2-10} 
 & word-LSTM & 2.2 & 5.86 &  36.83\% & -- & 49.19\%  & 41.44\% & -- & 32.41\% \\ \cline{2-10} 
 & elmo-LSTM & 2.38 & 5.87 & 37.91\% &  56\% & -- & 41.24\% & 44.37\% & -- \\ \hline
\multirow{3}{*}{IMDB} & char-LSTM & 11.15 & 19.62 & -- & 39.46\%&  34.69\%& -- & 22.28\% & 21.27\% \\ \cline{2-10} 
 & word-LSTM &  14.5& 19.79 & 31.33\% & -- &  40.56\%& 28.1\% & -- & 26.93\% \\ \cline{2-10} 
 & elmo-LSTM &  12.21& 19.74& 40.82\%& 53.06\%& -- & 41.49\% & 46.89\% & -- \\ \hline
\end{tabular}
}
\caption{AWR and Transferability of our adversaries in comparison to DWB. \textbf{AWR} : Average Words Replaced}
\label{tab:transfer-table}
\end{table*}


\subsection{Implementation Details}
\textbf{Classifiers:} We studied the efficacy of our algorithm on 3 classifiers that consume text in unique ways - a word-LSTM model (word-tokens), a char-LSTM model (character-tokens) and an elmo-LSTM model (word and character tokens). Every model was trained on two sentiment datasets viz. Stanford Sentiment Treebank (SST-2) \cite{socher-etal-2013-parsing} and IMDB movie reviews \cite{LMDB}. The char and word LSTMs were trained using the AllenNLP \cite{Gardner2017AllenNLP} toolkit. The elmo-LSTM was trained from scratch using a combination of Elmo \cite{Peters:2018} and GloVe embeddings.\\
\textbf{Sequence Mutator:} We trained $2$ different instances of the \emph{Sequence Mutator} - one for each dataset. In order to enhance its mutation capability, we train the mutator on a combined corpus of the test-set of the datasets and Wikitext-2 \cite{merity2016pointer}. This allowed the mutator to be finetuned for the particular dataset but yet at the same time be exposed to a general natural language understanding. To ensure predictions were accurate, we windowed long sentences in the dataset into short groups of utmost $30$ words. To ensure a fast inference, we limited the model size to $4.9*10^{7}$ parameters. While training, we used a learning rate of $10^{-3}$ and eventually stopped the process when the average softmax cross-entropy loss of the model reached $0.05$ or below. Table~\ref{table:sst_samples} lists few adversarial samples generated by our approach.

In order to validate the efficacy of our algorithm, we generate adversaries for test-set inputs in SST and IMDB, and record the drop in accuracies for each classifier. As a baseline, we also attacked the $3$ classifiers with the current state of the art approach DeepWordBug (\textit{DWB}) \cite{deepwordbug}.

\subsection{Evaluation}
\begin{figure}[ht]
\centering
\includegraphics[width=0.48\textwidth]{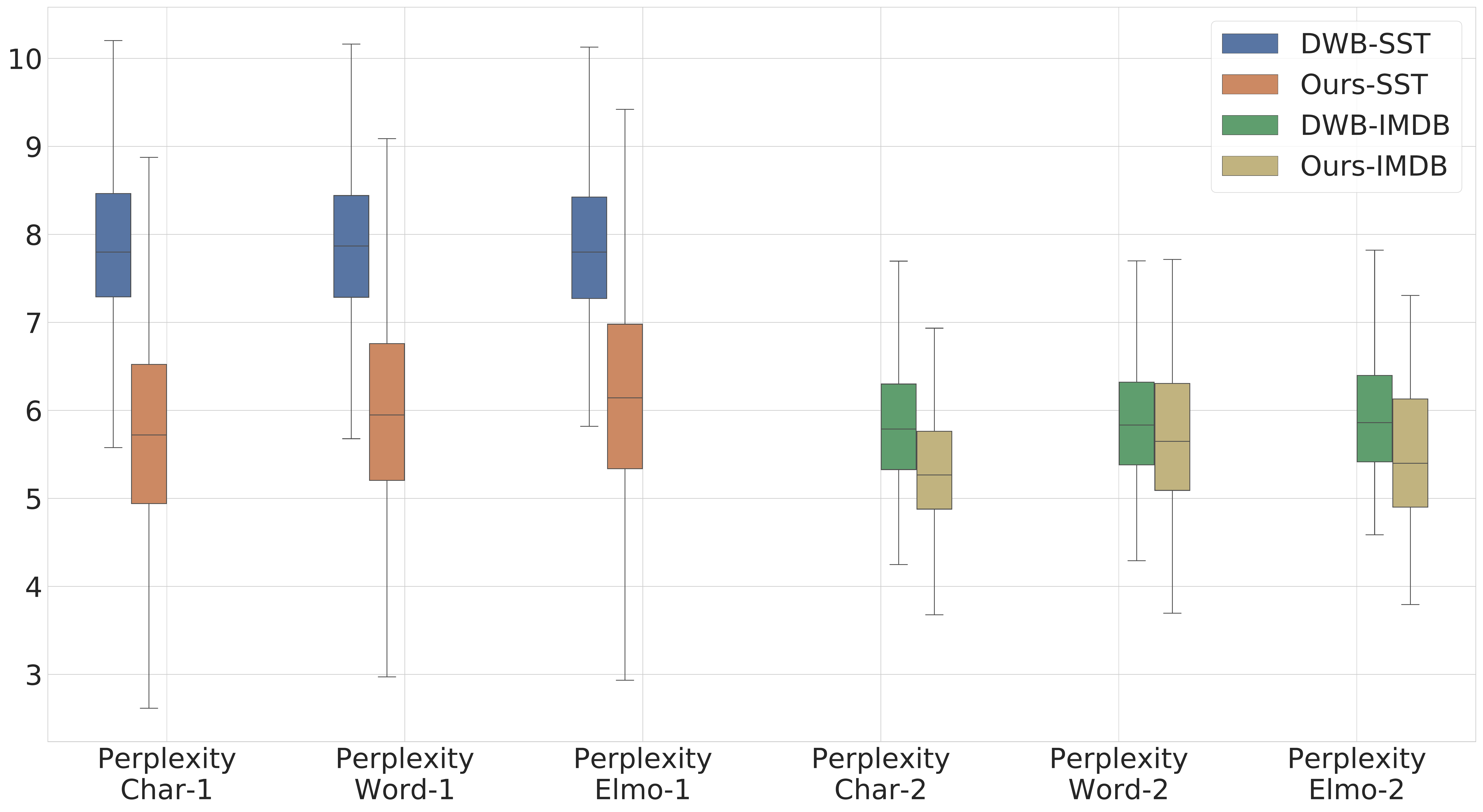}
\caption{Perplexity Score}
\label{fig:perplexity}
\end{figure}
We performed both quantitative (Table \ref{tab:my-table}) and qualitative (section \ref{section:quali-sec}) studies of our algorithm versus DWB. The study covers the following three dimensions - i) Performance degradation of the target models, ii) Transferability of  adversarial samples across models, iii) Desired adversarial qualities in the generated samples.


The effect of adversarial text on a model is measured by the \emph{success rate} and the decrease in model accuracy. Table \ref{tab:my-table} lists the original accuracy of the models, success rate of generated adversaries and the degraded accuracy of the models. There is a significant improvement over DWB across all the experiments. We also infer that our most successful strategy is the $Attack_{Combined}$ algorithm with a Sequence Mutator. On average, our success rate was $65.67\%$ for SST classifiers and $36.45\%$ for IMDB classifiers, where as, for DWB, it was only $43.93\%$ and $18.1\%$ respectively. As a result, there is a larger drop in accuracy when using our algorithm when compared to DWB as listed in Table~\ref{tab:my-table}. We also highlight the difference in success rates of the $Attack_{Single}$ mode with the Glove and Sequence mutator respectively. If one does not intend to allow \emph{typographical} adversaries, there is a significant benefit in using the Sequence mutator.

Further, Table~\ref{tab:transfer-table} shows the transferability of adversaries between the $3$ models. As shown, for the SST dataset, $41.44\%$ and $41.24\%$ of the adversaries curated for word-LSTM and elmo-LSTM respectively were transferable to char-LSTM. This clearly indicates that our attacks are transferable across models. As DWB aggressively introduces typographical errors in its adversaries, each typo maps to the unknown token irrespective of the classifier. Hence, as evident from Table~\ref{tab:transfer-table}, such adversaries are more transferable than those created by our approach. However, the increase in typos impact the quality of generated adversaries and affect its similarity with the original text. 

This is further verified in Fig. \ref{fig:reschunks} and Fig. \ref{fig:perplexity}, where we depict the distribution of various metrics that measure the quality of generated adversaries. We evaluate the adversaries using the objectives $f2$ and $f3$ along with natural language metrics like Blue, Meteor, Greedy matching score (GMS) and Vector extrema cosine similarity (VECS) \cite{sharma2017nlgeval}.  We obtain an average $f2$ value of $0.747$ for SST and $0.832$ for IMDB in comparison to DWB ($0.396$ and $0.756$ respectively). This $88.6\%$ increase for SST and $10.05\%$ increase for IMDB in $f2$ values indicates that we change far fewer words than DWB. This can be verified by the \emph{Average Words Replaced} statistic in Table \ref{tab:transfer-table}. 
We notice a similar trend for the $f3$ objective values as well. We also outperform DWB on word overlap metrics like Blue and Meteor. The Blue score for SST has a noticeably larger deviation for DWB when compared to our approach. Meteor mimics human judgment better than Blue by incorporating stemming and synonymy. We report higher Meteor scores in our generated adversaries signifying minimal perturbations. Similar improvements are noticed in sentence level metrics like VECS and GMS. VECS calculates the cosine similarity of sentence level embeddings whereas GMS calculates similarity scores  between   word embeddings and averages it for a sentence. Both of these sentence level metrics attain very high values with less deviation. These quantitative measures indicate that most of our adversaries enjoy high \emph{structural similarity} and \emph{semantic relevance}. In order to study the \emph{grammatical coherence} of our adversaries, we also study the $LM$ scores in Fig. \ref{fig:perplexity}. We observe that our average $LM$ score for generated adversaries in SST and IMDB were $5.94$ and $5.5$ which is $24.1\%$ and $5.66\%$ lower in comparison to DWB ($7.83$ and $5.83$) signifying better grammatical coherence. All the quantitative measures suggest that our algorithm performs well at maximizing adversarial qualities in all generated adversaries. However, this is further validated by a qualitative study.

\subsection{Qualitative Evaluation}
\label{section:quali-sec}
We conducted a human evaluation study to determine if the generated adversaries were in reality visually imperceptible to humans.
In order to assess the quality of the adversaries across various criteria, $34$ users (students and researchers from a computer science background) were presented with randomly sampled pairs of original and generated adversaries and then asked to answer multiple questions. We infer our insights from a total of $831$ data points obtained through this study.

Firstly, they were asked to predict the label of both the original and adversarial sample.  $94\%$ of the users identified the same label for both the original and adversarial text indicating that adversarial samples remained unperceivable to users but were yet able to fool the model. Secondly, the users were asked whether the pairs had the same semantic meaning. $85\%$ of users identified them as semantically similar. This proved that the adversarial samples retained enough details from the reference and were \textit{semantically relevant}. Finally, users were asked to rate the naturalness and grammatical correctness of each sample on a scale of $1-5$. The average score of $3.2$ for adversarial samples was very close to the score of $3.5$ for original samples which indicated that the adversarial samples were natural and grammatically coherent.

\section{Conclusion and Future Work}
We have successfully established that genetic algorithms can produce good quality adversaries if coupled with inference from deep learning based objectives and Seq2Seq mutation. Both the qualitative and quantitative evaluations indicate that most of the adversaries generated have desired adversarial qualities of \textit{structural similarity}, \textit{semantic relevance} and \emph{grammatical coherence}.

The possible future directions for this work are i) Experimentation with algorithm sub-modules - During Initialization, LIME \cite{lime} based strategy can be used to select important words for perturbation. ii) Data Augmentation - We can study the performance of the classifiers when trained with these additional adversarial samples. iii) Expanding Attacks - We can study our performance on tasks like question answering, summarization etc.

\bibliographystyle{acl_natbib}
\bibliography{emnlp2020}

\end{document}